\documentclass{article}

\usepackage[preprint]{arxiv_preprint}
\usepackage{amsmath}
\usepackage{amsfonts}
\usepackage{array}
\usepackage{booktabs}
\usepackage{graphicx}
\usepackage{placeins}
\usepackage{wrapfig}

\title{HORIZON: Recoverability-Governed Curriculum for Physical-Domain Scaling}

\author{
  \textbf{Chenhao Bai\textsuperscript{1},
  Liqin Lu\textsuperscript{2},
  Kaijun Wang\textsuperscript{1},
  Hui Chen\textsuperscript{1},
  Jin-Chuan Shi\textsuperscript{1},
  Yuyang Liu\textsuperscript{1}} \\
  \textbf{Hao Chen\textsuperscript{1,*},
  Chunhua Shen\textsuperscript{1,*}} \\
  \normalfont\textsuperscript{1}Zhejiang University, State Key Lab of CAD \& CG \\
  \textsuperscript{2}Zhejiang University of Technology \\
  \textsuperscript{*}Corresponding authors.
}

\begin{document}
\maketitle

\begin{figure}[!h]
    \centering
    \includegraphics[width=1.0\linewidth,height=0.6\linewidth]{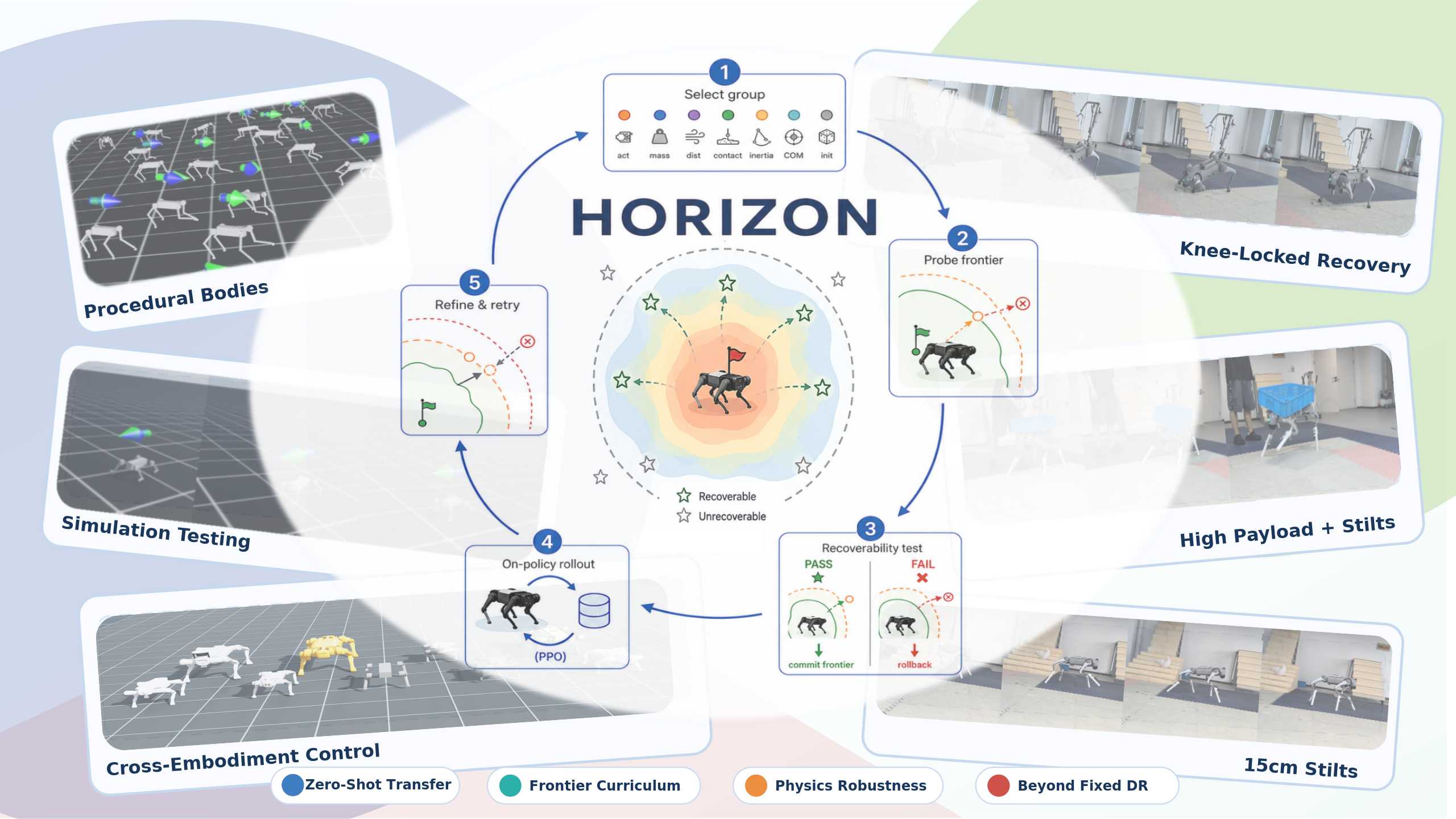}
    \caption{We present HORIZON, a recoverability-governed curriculum for physical-domain scaling in legged locomotion. HORIZON trains on generated robots and uses checkpointed rollback to expand physical domains only when harder dynamics still produce recoverable on-policy experience. This recoverability-gated training enables zero-shot transfer in simulation across diverse quadruped robots and, after hardware deployment, strong physical generalization under unseen real-world morphology and payload perturbations.}
    \label{fig:cover_deployment}
\end{figure}

\begin{abstract}
    Scaling robust robot policies requires more than broader randomization, because physical-domain experience must remain organized and learnable throughout training. We study when a policy can benefit from harder physics and identify recoverability as a central constraint in on-policy physical-domain scaling. In on-policy training, new dynamics are useful only insofar as they remain close enough to the current policy to generate corrective on-policy data, rather than collapsing rollouts into unrecoverable failures. Using quadruped locomotion as a physically demanding benchmark for embodied generalization, we introduce \textbf{HORIZON}, a checkpointed frontier curriculum that expands physical domains only within the current policy's recoverable boundary. HORIZON uses rollback and boundary refinement to govern each expansion step, turning fixed randomization into a continual process of physical-domain growth. Experiments reveal three regularities of physical-domain expansion. First, direct domain widening is uneven across physical axes and often unlearnable without staged ordering. Second, domain composition is non-monotonic, and adding more domains beyond a compact core can dilute recoverable joint samples and reduce overall robustness. Third, offline distillation of isolated experts cannot substitute for the joint interaction generated by on-policy curriculum. Together, these results frame physical-domain generalization as a continual growth problem for embodied control, with recoverability as the organizing principle for on-policy expansion.
\end{abstract}



\section{Introduction}
\label{sec:introduction}

Scaling laws have reshaped machine learning by showing that broader data and compute can yield smoother generalization trends \citep{kaplan2020scaling}. Robot learning motivates a related but more constrained question for physical control. In imitation learning, generalization can scale with environment and object diversity more strongly than with raw demonstration count \citep{lin2025data}. In online reinforcement learning, however, the relevant axis is not only data volume or model capacity, because data are produced by the current policy itself. The question is therefore how physical-domain experience can expand while remaining close enough to the current controller to be learnable.

This coupling turns physical-domain scaling into a cold-start problem for online robot control. A physical domain just beyond the current policy can expose useful corrective behavior, while one too far away produces short failures with little on-policy signal. Motors may lose torque or bandwidth, payloads may alter body dynamics, feet may enter new contact regimes, and disturbances may interact with balance before the controller can re-identify the system. These shifts change what actions are possible, which recovery behaviors are available, and how failures compound over time.

Quadruped locomotion is a useful benchmark for this question because it exposes many physical interfaces while still providing dense, measurable feedback. Unlike sparse long-horizon manipulation, locomotion failures appear quickly in reward, episode length, and terrain progress. At the same time, deployment-relevant variation spans actuation, mass, disturbances, contact, inertia, center-of-mass shift, and initial state. This makes locomotion a compact testbed for studying how physical-domain diversity should be organized in online RL.

Domain randomization is the standard way to expose policies to physical variation, but fixed ranges do not specify when a variation should enter training. If the range is too narrow, the policy learns a brittle controller around nominal physics. If it is widened too aggressively, on-policy RL may spend most updates in unrecoverable failures. Prior robust locomotion work observes that useful disturbances should push robots toward unstable but still recoverable states \citep{long2024hinfinity}; environment-design methods also emphasize informative training conditions \citep{jiang2021replay}; and controller parameters can act as learning interfaces rather than fixed settings \citep{bronars2026tune}. These views point to recoverability as a central constraint in on-policy physical-domain scaling.

We study this constraint through continual physical-domain expansion. For locomotion, recoverable frontier expansion introduces new dynamics as small enough changes from mastered behavior that the policy can preserve locomotion while extending where it can recover. We instantiate this idea in HORIZON, a checkpointed curriculum that organizes structured physical variation into learnable frontiers, uses rollback and boundary refinement to avoid unrecoverable expansion, and searches for compact core-domain compositions that improve composed out-of-distribution control.

This paper studies when additional physical experience helps an on-policy controller and reports three findings:
\begin{enumerate}
    \item Recoverability is an operational requirement for on-policy physical-domain expansion: harder dynamics must remain close enough to the current policy to produce corrective on-policy data, rather than collapsing rollouts into weak training signal.
    \item Physical-domain expansion has policy-dependent recoverability frontiers: different axes become learnable at different rates, and direct widening beyond the current frontier can collapse training rather than improve robustness.
    \item Physical diversity trades off against recoverable sample quality: compact core-domain curricula can be more effective than full-domain exposure, while offline distillation of isolated experts cannot replace joint on-policy interaction.
\end{enumerate}


\section{Related Work}
\label{sec:related_work}

\subsection{Domain Randomization and Sim2Real Transfer}
\label{subsec:related_dr}
Domain randomization improves sim2real robustness by training policies under randomized visual, dynamics, or contact parameters \citep{tobin2017domain,peng2018sim,tan2018sim}. Adaptive variants update the randomization distribution from real rollouts or informative samples \citep{chebotar2019closing,ramos2019bayessim,muratore2022npdr,mehta2020active}, but wider dynamics randomization is not always beneficial for quadruped locomotion \citep{xie2021dynamics}. HORIZON follows this line of work but expands physical ranges on-policy through committed, rolled-back, and refined frontiers.

\subsection{Curriculum Learning, Continual Reinforcement Learning, and Recoverable Frontiers}
\label{subsec:related_curriculum}
Curriculum learning orders tasks by difficulty, and automatic curricula select training conditions from learning progress or related signals \citep{bengio2009curriculum,narvekar2020curriculum,portelas2020teacher,florensa2017reverse}. Continual reinforcement learning studies how new tasks can be learned without erasing prior behavior, while RL's Razor links this retention to the KL-minimal update bias of online RL \citep{khetarpal2022continual,shenfeld2025rlrazor}. HORIZON uses checkpointed rollback and boundary refinement to make physical-domain expansion a recoverable curriculum rather than a fixed schedule.

\subsection{Generalization and Robust Motion Strategies}
\label{subsec:related_locomotion}
Robust quadruped locomotion has advanced through sim2real training, adaptation methods, perception-conditioned control, and generalist policies \citep{hwangbo2019learning,tan2018sim,rudin2022learning,kumar2021rma,lee2020learning,miki2022learning,liu2025locoformer,margolis2024rapid,zhuang2023robot,wang2026odyssey}. Most of these works improve the policy side through architectures, adaptation modules, perception interfaces, or training scale under a chosen randomization distribution. While LocoFormer scales across robot morphologies with aggressive domain randomization, HORIZON uses generated robot and physical-parameter variation to study the orthogonal question of how to organize physical-domain training so harder dynamics remain recoverable. This is complementary to embodiment-scale architectures and keeps the deployed actor interface unchanged.

\section{Method}
\label{sec:method}
This section defines the mechanism used to expand a locomotion policy beyond nominal physical conditions while keeping training recoverable. We first cast randomized physics as group-wise frontiers, then describe how the manager commits, retries, or rolls back candidate ranges before specifying the domain ranges. 

\begin{figure}[t]
    \centering
    \includegraphics[width=1.0\linewidth]{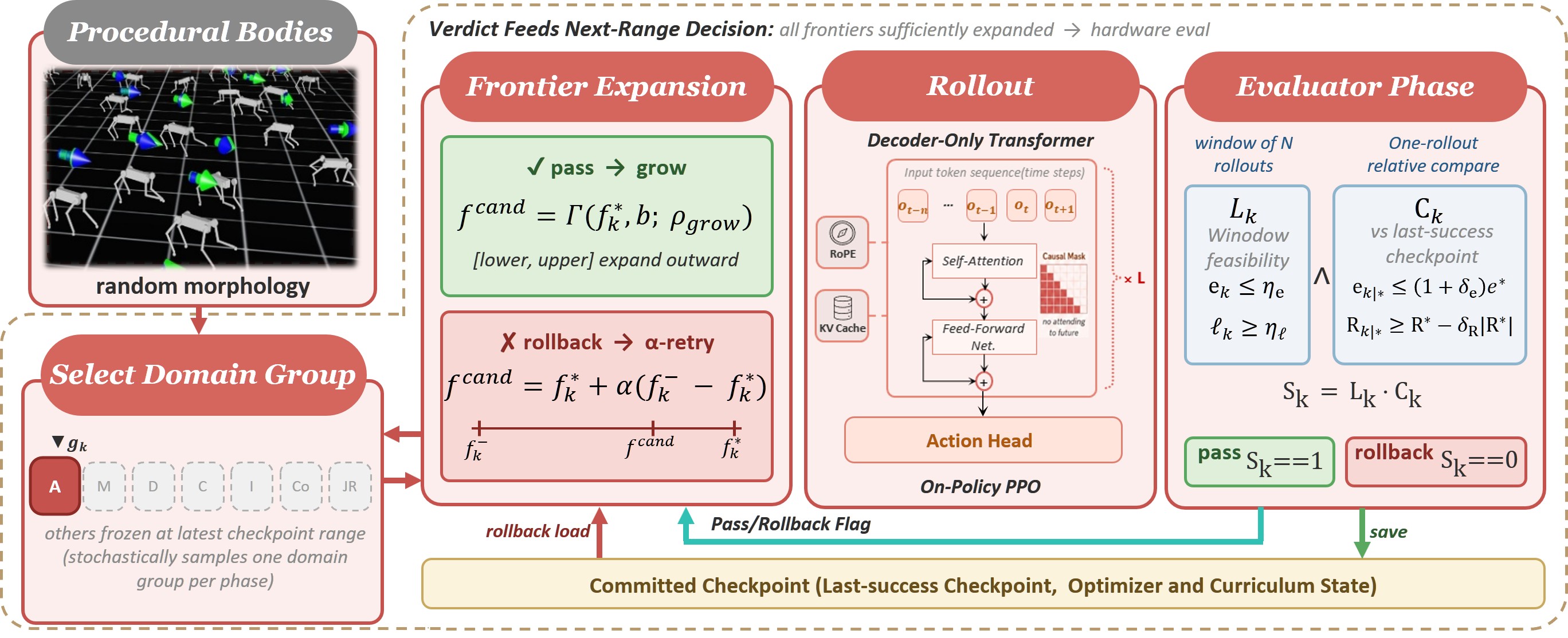}
    \caption{Overview of HORIZON recoverability-governed physical-domain scaling. Procedural robot bodies and a selected physical-domain group define the active frontier, while the remaining groups are frozen at the latest committed checkpoint to preserve failure attribution. For each candidate range, a decoder-only transformer policy is trained with on-policy PPO; implementation details are summarized in Tables~\ref{tab:policy_critic_settings} and~\ref{tab:ppo_settings}. The evaluator then combines windowed rollout feasibility with a one-rollout comparison against the last-success checkpoint to produce a pass-or-rollback verdict. Passing expands the mastered frontier and saves the optimizer and curriculum state, whereas failure records the failed range and curriculum state, reloads the committed checkpoint, and selects another domain group or nearby recoverable range for subsequent exploration.}
    \label{fig:method_overview}
\end{figure}

\subsection{Adaptive Domain Expansion}
\label{subsec:method_frontier_curriculum}
We treat physical conditions that exist in, or exceed, nominal real-world operation as a structured state space expanded during training. Instead of sampling from a fixed distribution, HORIZON searches near the current policy's learnable boundary, where dynamics expose weaknesses while remaining recoverable for on-policy learning. Each phase fixes a candidate physical range, collects rollouts, and then commits, rolls back, or refines the frontier:

\begin{equation}
    (g_k,\xi_k)=\mathcal{M}(c_k),\quad
    \tau_k\sim p_{\theta,\xi_k},\quad
    c_{k+1}=\mathcal{U}\!\left(c_k,g_k,\xi_k,\mathcal{E}(\tau_k,c_k)\right),
    \qquad \xi_k\in\Omega_{g_k}.
    \label{eq:phase_update}
\end{equation}

Here $c_k$ stores each group's mastered frontier, failed bracket, pending retry, and refine state. The manager $\mathcal{M}$ selects group $g_k$ and range $\xi_k\in\Omega_{g_k}$; $\tau_k$ are rollouts from policy-simulator distribution $p_{\theta,\xi_k}$; $\mathcal{E}$ evaluates them; and $\mathcal{U}$ updates the curriculum. A pass advances the mastered frontier, while failure tightens the bracket, creates a retry, or, at the lowest refine rung, marks a local learnability boundary.

\subsection{Frontier States and Phase Updates}
\label{subsec:method_phase_updates}

For each domain group, the manager maintains mastered, candidate, failed, and pending-retry frontiers. At phase end, the evaluator passes a candidate only if both the locomotion feasibility gate $\mathcal{L}_k$ and checkpoint-comparison gate $\mathcal{C}_k$ pass:
\begin{equation}
    \mathcal{L}_k =
    \mathbb{I}\!\left[
    \ell_k \ge \eta_{\ell},\;
    e_k \le \eta_e
    \right].
    \label{eq:locomotion_gate}
\end{equation}
\begin{equation}
    \mathcal{C}_k =
    \mathbb{I}\!\left[
    e_{k\mid *} \le (1+\delta_e)e^{*},\;
    R_{k\mid *} \ge R^{*}-\delta_R|R^{*}|
    \right].
    \label{eq:checkpoint_compare_gate}
\end{equation}
\begin{equation}
    \mathcal{S}_k = \mathcal{L}_k \cdot \mathcal{C}_k .
    \label{eq:success_gate}
\end{equation}
The binary signal $\mathcal{S}_k$ is therefore a finite-horizon recoverability gate. In $\mathcal{L}_k$, $\ell_k$ is candidate-frontier episode length, $e_k$ is tracking error, and $\eta_{\ell},\eta_e$ are the minimum length and maximum error thresholds. In $\mathcal{C}_k$, $e_{k\mid *}$ and $R_{k\mid *}$ are the current policy's tracking error and return in the checkpoint environment; $e^{*}$ and $R^{*}$ are checkpoint statistics; and $\delta_e,\delta_R$ are tolerance margins. Candidate feasibility is necessary but not sufficient because checkpoint behavior must also be preserved; Table~\ref{tab:recoverability_predictive_diagnostic} shows that this comparison adds predictive signal beyond rollout feasibility. Settings and the phase-success versus fixed-OOD distinction are in Sec.~\ref{app:curriculum_manager_success}.

For an active frontier coordinate, the manager proposes the next challenge by

\begin{equation}
    f_{k+1}^{\mathrm{cand}}
    =
    \begin{cases}
        \Gamma(f_k^{*}, b; \rho_{\mathrm{grow}}),
        & \text{coarse proposal},\\
        f_k^{*}+\alpha(f_k^{-}-f_k^{*}),
        & \text{recovery proposal},\\
    \end{cases}
    \qquad 0 < \alpha < 1 ,
    \label{eq:frontier_proposal}
\end{equation}
Here $f_k^{*}$ is the mastered checkpoint frontier, $f_k^{-}$ is the failed frontier, $b$ is the curriculum limit, $\Gamma$ expands toward that limit with growth factor $\rho_{\mathrm{grow}}$, and $\alpha$ is the recovery interpolation weight. Thus coarse proposals explore outward, recovery proposals retreat between the checkpoint and failed frontier, and the resulting pass/fail signal either commits a new checkpoint or tightens or terminates that search direction.

\subsection{Domain Groups and Curriculum Ranges}
\label{subsec:method_domain_groups}

In designing the randomization domains, we do not restrict them to sim2real failure modes. Instead, we factorize a broader physical state space into action authority, morphology scale, external perturbations, contact conditions, body inertia, COM shift, and initial state. The complete baseline ranges and curriculum limits are reported in Table~\ref{tab:domain_ranges}. The baseline ranges correspond to normalized randomization \citep{long2024hybrid}, while the curriculum limits provide the manager with a wider search region for gradually identifying learnable physical boundaries that exist in, or exceed, nominal real world operation. In focused mode, only one domain group is advanced at a time, while the others remain at the latest checkpoint range to preserve failure attribution.


\section{Experimental Results and Analysis}
\label{sec:experiments}

This section tests whether recoverable frontier expansion can sustain physical-domain growth and identify compact domain compositions for generalization. Unless otherwise stated, success denotes the stacked seven-domain \texttt{OOD all} evaluation, a deliberately beyond-realistic stress test where all physical shifts exceed nominal real-world operating ranges.

\subsection{Evaluation Protocol}
\label{subsecresprotocolrollback}

\textbf{Rollback tests recoverability.} The stability ablation isolates rollback by comparing HORIZON with a \texttt{w/o rollback} variant and a \texttt{Direct wide DR} baseline that exposes the policy to the target hard ranges from the start.

\begin{wrapfigure}{r}{0.44\textwidth}
    \vspace{-1.0em}
    \centering
    \includegraphics[width=\linewidth]{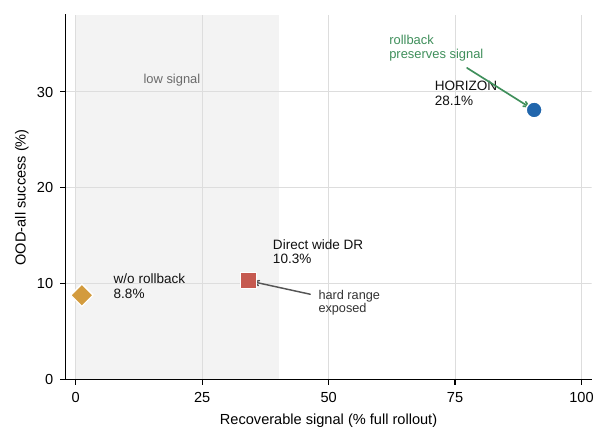}
    \caption{Plateau-checkpoint recoverable signal versus \texttt{OOD all} success.}
    \label{figablation}
    \vspace{-1.0em}
\end{wrapfigure}

The evaluation in Figure~\ref{figablation} uses plateaued checkpoints for each run rather than a shared fixed iteration count. The horizontal axis reports the recoverable rollout signal, normalized as the percentage of a full episode completed at the evaluated checkpoint. At these checkpoints, both failed baselines remain at low rollout signal. Without rollback, the policy cannot reliably remain upright under the hard range, while direct wide-DR training exposes the target randomization from the start and collapses into unreliable rollouts. HORIZON keeps episode length and \texttt{OOD all} success higher, indicating that hard-range exposure helps only when rollouts remain recoverable enough to provide corrective on-policy data. A diagnostic over 1,509 logged frontier phases further shows that the composite gate predicts subsequent frontier progress better than episode length alone (Sec.~\ref{app:recoverability_diagnostic}).

\FloatBarrier

\subsection{Recoverability Governs Physical-Domain Expansion}
\label{subsecressingledomain}

\begin{figure}[!htbp]
    \centering
    \includegraphics[width=1.0\linewidth,height=0.32\linewidth]{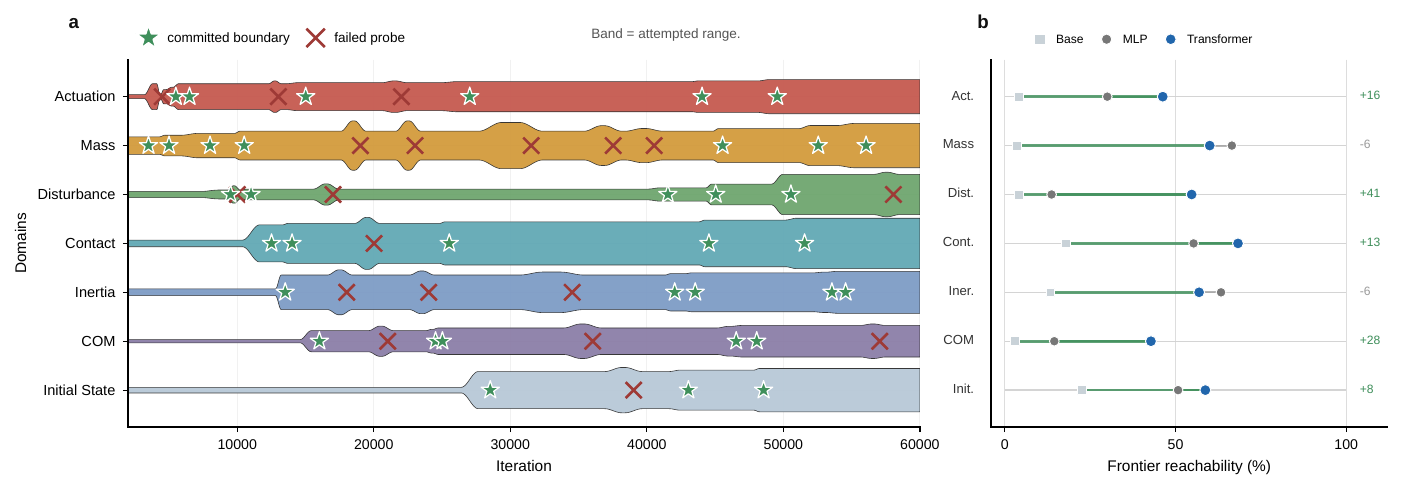}
    \caption{Recoverability-governed physical-domain expansion in full-domain training. HORIZON sequentially unlocks and expands domain groups rather than enabling all DR groups from the start. Panel a visualizes the transformer run, and panel b compares frontier reachability with base randomization and a full-domain MLP curriculum.}
    \label{figsingledomain}
\end{figure}

\textbf{Expansion succeeds only where failure remains recoverable.} The rollback ablation shows that hard physical ranges cannot simply be enabled at once, but leaves open which axes can later absorb failed expansion. We diagnose this by treating each group as a sequence of frontier probes and asking whether wider ranges become commit-able after recovery. In Figure~\ref{figsingledomain}, colored bands denote attempted outward moves, crosses mark rolled-back probes, and stars mark committed boundaries.

Failed probes can be productive rather than terminal. Actuation fails at about $4.5$k iterations, yet later commits a substantially wider boundary after intermediate recovery phases; disturbance and COM show the same pattern. Base randomization covers only small fractions of these hard frontiers, including $4.1\%$ for actuation, $4.0\%$ for disturbance, and $3.0\%$ for COM. HORIZON raises the corresponding coverage to $46.3\%$, $54.7\%$, and $42.8\%$, exceeding the MLP comparison by $16$, $41$, and $28$ percentage points.

This analysis separates recoverable bottlenecks from axes that are already easy or saturated within full-domain training. As shown in Fig.~\ref{fig:domain_composition}a, some groups expand with strong per-axis generalization, whereas the full domain set spends curriculum budget on axes that may provide little additional transfer. The next section therefore asks whether smaller coupled groups can retain useful recovery transfer at lower learning cost.

\FloatBarrier

\subsection{\texorpdfstring{Core-Domain Composition}{Core-Domain Composition}}
\label{subsec:res_core_domain}

\textbf{Multi-domain physical scaling is a relation problem, not a coverage problem.} We search for domain relations that remain jointly recoverable under coupled perturbations, seeking a compact core that creates shared recovery demands without spending phase budget on easy, saturated, or interfering groups.

\begin{figure}[t]
    \centering
    \includegraphics[width=0.95\linewidth]{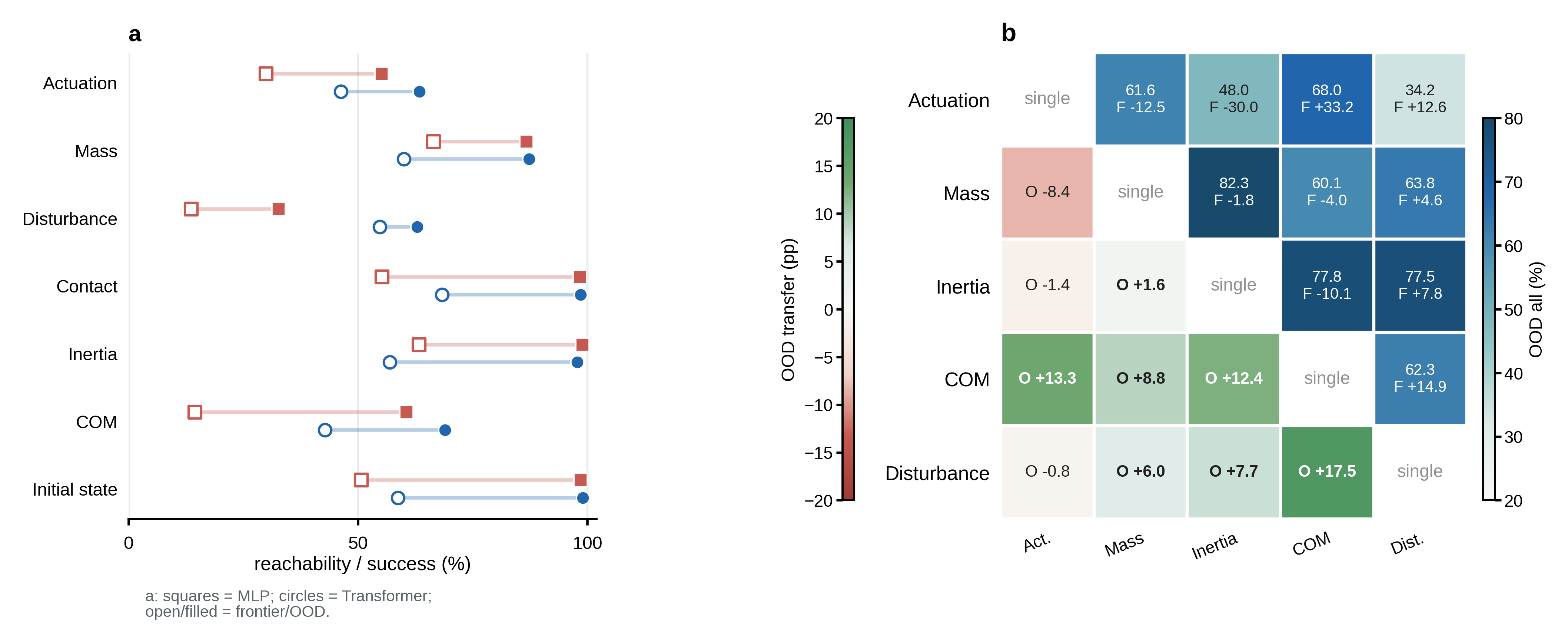}
    \caption{Composition-search diagnostic. Panel a contrasts full-domain frontier reachability with per-group OOD success for the MLP and transformer policies. Panel b reports pairwise two-domain \texttt{OOD all}, frontier change $F$, and isolated transfer $O$.}
    \label{fig:domain_composition}
\end{figure}

\begin{wrapfigure}[12]{l}{0.50\textwidth}
    \vspace{-0.8em}
    \centering
    \includegraphics[width=\linewidth]{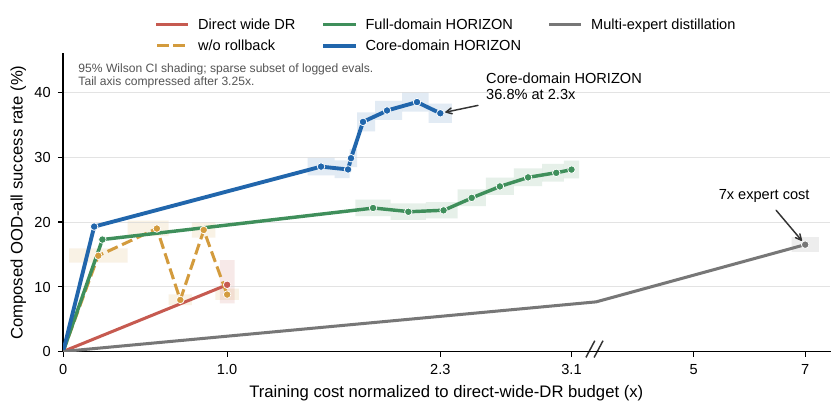}
    \caption{Cost-normalized composed \texttt{OOD all} during curriculum training.}
    \label{fig:sample_efficiency}
    \vspace{0.8em}
\end{wrapfigure}

This criterion is supported by the per-group diagnostic in Fig.~\ref{fig:domain_composition}a, where contact, inertia, and initial state maintain high per-group OOD success even with moderate frontier reachability. Widening every group is therefore not the main bottleneck. The harder question is whether coupled perturbations still generate recoverable rollouts, and the pairwise matrix makes this relation structure explicit. \texttt{Actuation + COM} raises composed success to $68.0\%$ while improving the hard actuation bottleneck, and \texttt{COM + disturbance} improves isolated transfer while expanding the mastered frontier. Other pairs add diversity but reduce transfer or frontier coverage, suggesting conflicting failure samples. The matrix therefore screens for hard axes that still need coupled recovery pressure and partners that improve transfer, leading to the compact actuation, mass, disturbance, and COM core.

The cost-normalized view in Figure~\ref{fig:sample_efficiency} turns this relation structure into the main physical-domain growth result. The four-domain core is more efficient than full-domain HORIZON and multi-expert distillation, reaching stronger composed \texttt{OOD all} with less training cost. Direct wide DR and the no-rollback variant remain weak under hard joint physics, while full-domain training spends phases on axes already strong in isolation. The compact core concentrates updates on action authority, payload shift, external recovery, and asymmetric loading, preserving dense recoverable samples for learning physical interactions.

\subsection{Comparative Generalization Results}
\label{subsec:res_expert_distillation}

Table~\ref{tab:core_domain_results} compares the selected four-domain core with full-domain curriculum training, multi-expert distillation, ADR, and GRAM under the same fixed-OOD suite, whose hardest column, \texttt{OOD all}, shifts all seven physical groups together beyond realistic ranges. Baseline details are given in Sec.~\ref{app:baseline_comparison_details}.

\begin{table}[hbt]
    \centering
    \caption{OOD success rates (\%). Values are mean $\pm$ standard deviation over five trained seeds, each evaluated with the same fixed-OOD protocol.}
    \label{tab:core_domain_results}
    \resizebox{\linewidth}{!}{%
    \begin{tabular}{lcccccccc}
        \toprule
        Training setting & Actuation & Mass & Disturbance & Initial state & Contact & Inertia & COM & 7-domain OOD all \\
        \midrule
        Core-domain 
        & $\mathbf{69.3 \pm 2.8}$ 
        & $\mathbf{92.1 \pm 1.1}$
        & $\mathbf{68.0 \pm 2.0}$
        & $92.0 \pm 1.4$
        & $94.5 \pm 0.8$
        & $91.0 \pm 1.1$
        & $\mathbf{71.1 \pm 1.8}$
        & $\mathbf{36.8 \pm 2.5}$ \\
        Full-domain 
        & $63.4 \pm 2.5$
        & $87.3 \pm 1.2$
        & $62.9 \pm 1.2$
        & $\mathbf{99.0 \pm 0.5}$
        & $\mathbf{98.5 \pm 0.7}$
        & $\mathbf{97.8 \pm 0.9}$
        & $69.0 \pm 2.5$
        & $28.1 \pm 1.8$ \\
        Multi-expert distillation \citep{rusu2016policy}
        & $45.4 \pm 2.2$
        & $53.9 \pm 1.7$
        & $26.4 \pm 2.1$
        & $80.9 \pm 1.5$
        & $78.0 \pm 2.0$
        & $81.4 \pm 1.0$
        & $49.1 \pm 1.2$
        & $16.5 \pm 1.2$ \\
        GRAM \citep{queeney2025gram}
        & $60.7 \pm 1.3$
        & $69.4 \pm 2.0$
        & $37.9 \pm 0.2$
        & $70.0 \pm 1.8$
        & $70.0 \pm 1.4$
        & $71.2 \pm 1.7$
        & $51.0 \pm 1.0$
        & $22.7 \pm 1.1$ \\
        ADR \citep{mehta2020active}
        & $50.0 \pm 2.1$
        & $47.8 \pm 2.3$
        & $9.4 \pm 1.3$
        & $54.7 \pm 2.4$
        & $54.7 \pm 1.1$
        & $57.6 \pm 0.6$
        & $34.9 \pm 0.4$
        & $14.7 \pm 0.7$ \\
        \bottomrule
    \end{tabular}%
    }
\end{table}

The core curriculum obtains the strongest seven-domain \texttt{OOD all} result, improving full-domain training by $8.7$ points on the composed stress while remaining competitive on individual axes. Full-domain training is strongest on easier isolated groups but weaker under simultaneous perturbations, suggesting that adding domains can dilute recoverable joint samples. Multi-expert distillation and ADR perform substantially worse, and GRAM is competitive but below the core. Additional checks in Table~\ref{tab:heldout_core_results} show the same scope, with the compact core strongest on the target stress and nearby extensions, while full-domain training is slightly better on a few high-success non-core-heavy or contact-heavy compositions.

\subsection{Real-World Evaluation}
\label{subsec:res_real_world_eval}

\textbf{Deployment tests target morphology and skill transfer.} We evaluate the final policy zero-shot on hardware interventions that instantiate the same physical axes studied in simulation, but are not replayed as calibrated simulator parameters during deployment. The real-world suite is designed as a set of difficult deployment tasks over joint constraints, leg-length changes, asymmetric morphology, payload shifts, and coupled morphology-payload changes. These tests compare the HORIZON policy against a base-DR baseline trained with the same deployment interface, checking whether recoverable physical-domain expansion transfers to concrete hardware perturbations without policy fine-tuning or additional deployment inputs. Beyond the Go2 hardware suite, the final trained model also supports cross-embodiment locomotion control in simulation across additional quadruped bodies, with representative rollouts shown in Fig.~\ref{fig:cross_robot_control_montage}. Sec.~\ref{app:real_world_hardware_details} gives the evaluation protocol and baseline outcomes.

\begin{figure}[hbt]
    \centering
    \includegraphics[width=1.0\linewidth]{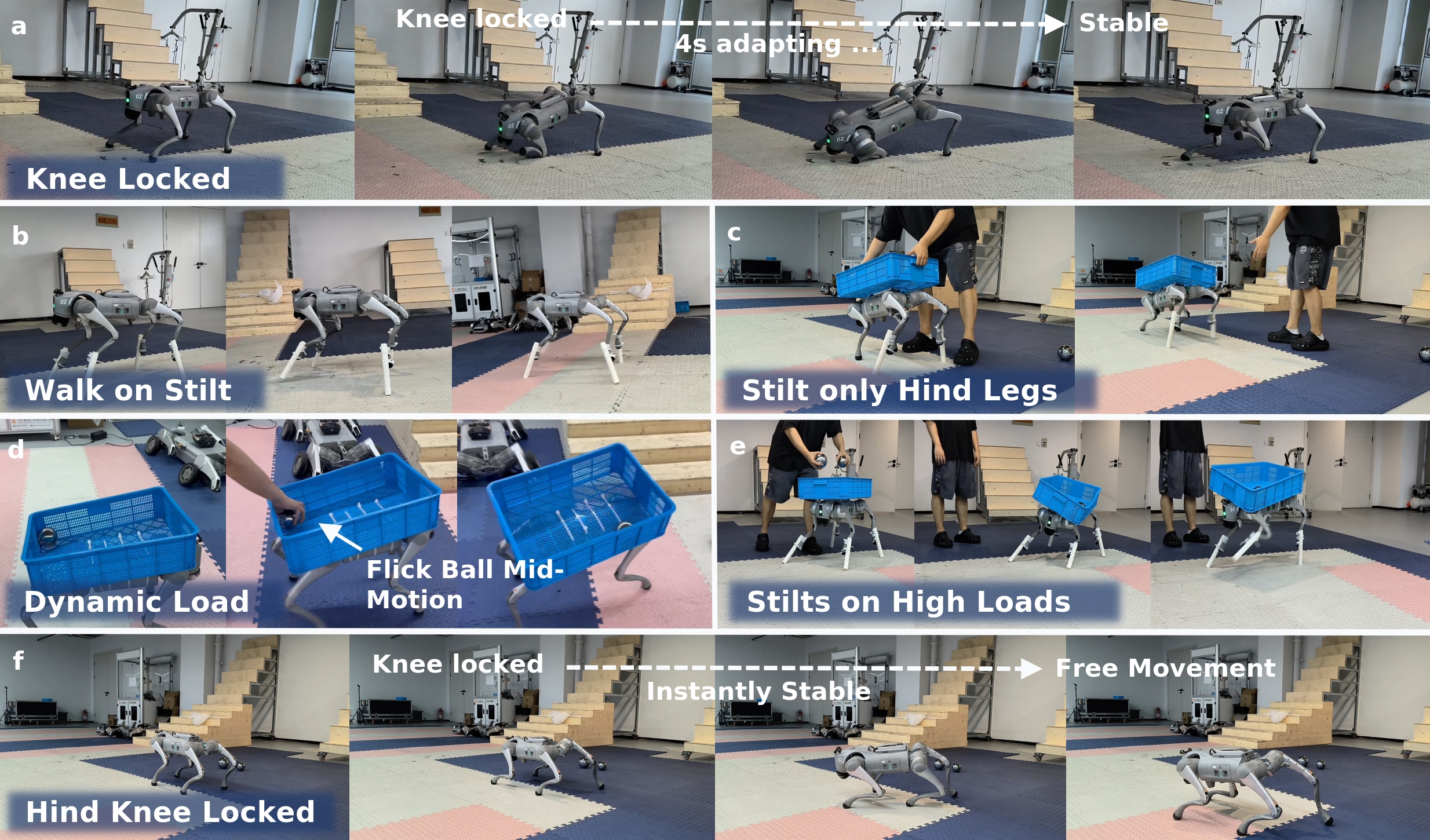}
    \caption{Real-world evaluation setup. These unseen OOD hardware tests stress morphology and payload transfer through hind-knee-locked locomotion, stilt walking, hind-leg-only stilts, dynamic loading, and high-load stilted locomotion.}
    \label{fig:real_world_eval}
\end{figure}
\FloatBarrier


\section{Conclusion}
\label{sec:discussion}

Recoverability organizes physical-domain scaling in on-policy robot RL. New dynamics help only while they remain close enough to the current policy to produce corrective rollouts, linking HORIZON to the local update bias emphasized by RL's Razor \citep{shenfeld2025rlrazor}. The experiments show three scaling regularities. Direct widening is uneven across physical axes, domain composition is non-monotonic, and behavior compression cannot replace composed on-policy interaction. As in imitation learning, diversity can matter more than quantity \citep{lin2025data}, but physical-domain RL further requires diversity to stay recoverable. The actuation, mass, disturbance, and COM core therefore outperforms full seven-domain exposure on the target composed stress while remaining competitive across held-out compositions, preserving recoverable sample density rather than maximizing nominal coverage.

\section{Limitations}
\label{sec:limitations}

HORIZON is currently a locomotion curriculum over a hand-designed physical-domain space. Extending this view to loco-manipulation will require curricula that coordinate body dynamics, contact-rich object interaction, and task progress, because recoverability may depend on both maintaining mobility and preserving manipulation affordances. A second limitation is that the domain groups are specified manually. Language-model-assisted domain discovery could generate candidate physical-domain definitions from robot morphology, task descriptions, and failure traces, but those candidates would still need recoverability tests before entering on-policy training. Finally, this work studies physical-domain variation rather than visually conditioned control. The same principle may have an analogue in perception, where appearance, lighting, geometry, and occlusion shifts should expand only when they remain recoverable for the current policy. Connecting perceptual-domain and physical-domain curricula is therefore a natural next step.



\bibliography{example}  

\clearpage
\appendix

\section{Recoverability Diagnostic}
\label{app:recoverability_diagnostic}

The phase gate in Eqs.~\ref{eq:locomotion_gate}--\ref{eq:success_gate} is used only to decide whether a candidate frontier should be committed during training. It is not used as the reported fixed-OOD metric, and we do not interpret it as an exact viability kernel. To check whether the gate measures more than current rollout length, we retrospectively analyzed 1,509 logged frontier phases from 65 existing curriculum runs. For each phase, we computed the subsequent change in the logged normalized mastered-frontier difficulty over the next three phase decisions. This target is a training-progress quantity rather than the final OOD success score, reducing direct overlap with the evaluation metric.

\begin{table}[hbt]
    \centering
    \setlength{\tabcolsep}{3pt}
    \caption{Predictive diagnostic for the operational recoverability gate. ``Length pass, checkpoint fail'' phases have high current episode length but fail the checkpoint-comparison gate, separating survival from recoverable training progress.}
    \label{tab:recoverability_predictive_diagnostic}
    \resizebox{\linewidth}{!}{%
    \begin{tabular}{@{}p{0.28\linewidth}ccccc@{}}
        \toprule
        Frontier class & Phases & Episode fraction & Checkpoint reward ratio & $\Delta$ frontier difficulty & Improvement rate \\
        \midrule
        Full gate pass
        & 656 & 0.865 & 1.039 & $+0.0635$ & 85.8\% \\
        Length and tracking pass, checkpoint fail
        & 414 & 0.891 & 0.775 & $+0.0285$ & 55.6\% \\
        Low rollout length
        & 433 & 0.682 & 0.387 & $+0.0100$ & 16.2\% \\
        \bottomrule
    \end{tabular}
    }
\end{table}

The diagnostic shows that long episodes are not sufficient evidence of recoverability. The length-only failure group has slightly higher current episode fraction than the full-pass group, yet produces less than half the subsequent frontier gain and a much lower improvement rate. Across the same phase records, Spearman correlation with future frontier gain is 0.35 for episode fraction and 0.52 for the composite recoverability score that includes checkpoint preservation. We therefore use recoverability in the paper as a predictive, finite-horizon training signal: it estimates whether a frontier is close enough to the checkpoint policy to provide useful corrective on-policy data, while final OOD success remains a separate fixed-evaluation endpoint.

\section{Training Details}
\label{app:training_details}

\subsection{Reward Function}
\label{app:reward_function}

Table~\ref{tab:reward_terms} lists the active reward terms and weights; disabled or zero-weight terms do not contribute to the return.

\begin{table}[hbt]
    \centering
    \caption{Reward terms and weights.}
    \label{tab:reward_terms}
    \begin{tabular}{@{}p{0.30\linewidth}p{0.47\linewidth}p{0.15\linewidth}@{}}
        \toprule
        Name & Term $\phi_i$ & $w_i$ \\
        \midrule
        Linear velocity
        & $\exp(-\lVert c_{xy}-v_{xy}\rVert_2^2 / \sigma^2)$
        & $1.25$ \\
        Yaw velocity
        & $\exp(-(c_{\omega_z}-\omega_z)^2 / \sigma^2)$
        & $1.25$ \\
        Vertical velocity
        & $v_z^2$
        & $-2.0$ \\
        Roll/pitch angular velocity
        & $\lVert \omega_{xy}\rVert_2^2$
        & $-0.05$ \\
        Action rate
        & $\lVert a_t-a_{t-1}\rVert_2^2$
        & $-0.02$ \\
        Feet air time
        & $\sum_j (T^{\mathrm{air}}_j-0.5)$
        & $0.2$ \\
        Flat orientation
        & $\lVert g^{\mathrm{proj}}_{xy}\rVert_2^2$
        & $-2.5$ \\
        Hip deviation
        & $\lVert q-q_0\rVert_1$
        & $-0.4$ \\
        Thigh/calf deviation
        & $\lVert q-q_0\rVert_1$
        & $-0.04$ \\
        Adaptive base height
        & $(z-z^\star)^2$
        & $-5.0$ \\
        Normalized action smoothness
        & $\lVert \bar{a}_t-\bar{a}_{t-1}\rVert_2$
        & $-0.02$ \\
        Normalized joint power
        & $\sum_j \lvert \tau_j\dot{q}_j\rvert / m$
        & $-2.0{\times}10^{-5}$ \\
        Normalized torque
        & $\lVert \bar{\tau}\rVert_2^2$
        & $-2.0{\times}10^{-4}$ \\
        Normalized joint acceleration
        & $\lVert \bar{\ddot{q}}\rVert_2^2$
        & $-2.5{\times}10^{-7}$ \\
        \bottomrule
\end{tabular}
\end{table}

\subsection{Algorithm Settings}
\label{app:algorithm_settings}

The policy is a decoder-only transformer conditioned on recent proprioceptive observations, commands, and previous actions. The simulation environments are implemented in Isaac Lab \citep{mittal2025isaaclab}, and we train with PPO from \texttt{rsl\_rl} \citep{schwarke2025rslrl}, using causal attention, RoPE, and a sliding-window key-value cache from \texttt{cusrl} \citep{zhu2024safe}. The actor receives only deployment-available proprioception, while simulation-only privileged signals enter the critic MLP. During training, cached historical inputs are recomputed, old key-value representations are detached across resets, and hidden states are cleared on environment reset, curriculum phase changes, or forced recovery. Tables~\ref{tab:policy_critic_settings} and~\ref{tab:ppo_settings} summarize the network and PPO settings used for training.

\begin{table}[hbt]
    \centering
    \caption{Policy and critic settings.}
    \label{tab:policy_critic_settings}
    \begin{tabular}{@{}p{0.38\linewidth}p{0.54\linewidth}@{}}
        \toprule
        Parameter & Value \\
        \midrule
        Policy class & decoder-only transformer \\
        Initial action noise std & $1.2$ \\
        Activation & GELU \\
        Embedding dimension & $128$ \\
        Attention heads & $8$ \\
        Feedforward dimension & $512$ \\
        Context window & $127$ \\
        RoPE base & $10000$ \\
        Critic hidden dimensions & $[512,256,128]$ \\
        Privileged encoder hidden dimension & $64$ \\
        \bottomrule
    \end{tabular}
\end{table}

\begin{table}[hbt]
    \centering
    \caption{PPO settings.}
    \label{tab:ppo_settings}
    \begin{tabular}{@{}p{0.38\linewidth}p{0.54\linewidth}@{}}
        \toprule
        Parameter & Value \\
        \midrule
        Steps per environment & $24$ \\
        Value loss coefficient & $1.0$ \\
        Clipped value loss & enabled \\
        Clip range & $0.2$ \\
        Entropy coefficient & $0.01$ \\
        Learning epochs & $8$ \\
        Minibatches & $4$ \\
        Learning rate & $5{\times}10^{-5}$ \\
        Learning-rate schedule & adaptive \\
        Discount factor $\gamma$ & $0.99$ \\
        GAE $\lambda$ & $0.95$ \\
        Target KL & $0.01$ \\
        Max gradient norm & $1.0$ \\
        \bottomrule
    \end{tabular}
\end{table}

\subsection{Curriculum Manager and Success Criteria}
\label{app:curriculum_manager_success}

The curriculum manager uses a small set of hyperparameters to separate frontier proposal, training-time evaluation, and fixed-OOD reporting. Table~\ref{tab:curriculum_manager_success_settings} groups the most important settings into curriculum dynamics, evaluation criteria used for phase commits, and OOD success metrics.

\begin{table}[hbt]
    \centering
    \scriptsize
    \setlength{\tabcolsep}{3pt}
    \renewcommand{\arraystretch}{0.88}
    \caption{Key thresholds used by HORIZON. The first row controls recovery proposals after failed frontiers; the middle rows define the training-time commit gate; the last row defines fixed-OOD episode success.}
    \label{tab:curriculum_manager_success_settings}
    \resizebox{\linewidth}{!}{%
    \begin{tabular}{@{}p{0.34\linewidth}p{0.22\linewidth}p{0.38\linewidth}@{}}
        \toprule
        Setting / criterion & Value & Purpose \\
        \midrule
        Recovery interpolation $\alpha$
        & $0.4$
        & Interpolate from mastered to failed frontier. \\
        Phase window
        & $500$
        & Recent rollout window. \\
        Length gate $\ell_k$
        & $0.85L_{\max}$
        & Minimum phase rollout length. \\
        Tracking gate $e_k$
        & $0.5$
        & Maximum phase tracking error. \\
        Checkpoint tracking gate $e_{k\mid *}$
        & $1.05$
        & Allowed tracking degradation. \\
        Checkpoint reward gate $R_{k\mid *}$
        & $0.05$
        & Allowed reward degradation. \\
        OOD episode success
        & $0.95L_{\max},\ 0.4$
        & Length and tracking thresholds. \\
        \bottomrule
    \end{tabular}
    }
\end{table}

During training, the evaluation gates decide whether a candidate frontier is committed. During fixed-OOD evaluation, success is computed under frozen OOD ranges using the episode-level criterion in the last row.

\subsection{Randomization Domains}
\label{app:domain_randomization_ranges}

Table~\ref{tab:domain_ranges} lists the baseline ranges used for near-nominal randomization and the curriculum limits used by the manager when expanding each physical-domain group.

\begin{table}[hbt]
    \centering
    \setlength{\tabcolsep}{3pt}
    \caption{Domain Randomization design. Baseline ranges define near-nominal randomization; curriculum limits define the manager's search region.}
    \label{tab:domain_ranges}
    \begin{tabular}{@{}>{\raggedright\arraybackslash}p{0.15\linewidth}>{\raggedright\arraybackslash}p{0.28\linewidth}>{\raggedright\arraybackslash}p{0.25\linewidth}>{\raggedright\arraybackslash}p{0.25\linewidth}@{}}
        \toprule
        Group & Parameters & Baseline range & Curriculum limit \\
        \midrule
        Actuation
        & effort, velocity, stiffness, damping
        & $[39,41]$, $[29,31]$, $[39,41]$, $[0.9,1.1]$
        & $[20,80]$, $[10,60]$, $[20,80]$, $[0.4,4.0]$ \\
        Mass
        & rigid-body mass scale
        & $[0.9,1.1]$
        & $[0.4,5.0]$ \\
        Disturbance
        & push velocity, external force, external torque
        & $[-0.5,0.5]$, $[-0.02,0.02]$, $0.0$
        & $[-5.0,5.0]$, $[-1.0,1.0]$, $[-0.5,0.5]$ \\
        Contact
        & static/dynamic friction, restitution
        & $[0.4,2.0]$, $[0.4,2.0]$, $0.0$
        & $[0.05,6.0]$, $[0.05,6.0]$, $1.0$ \\
        Inertia
        & inertia scale
        & $[0.9,1.1]$
        & $[0.5,2.0]$ \\
        COM
        & base COM offset
        & $[-0.003,0.003]$
        & $[-0.3,0.3]$ \\
        Joint and reset
        & initial joint position/velocity
        & $[0.9,1.1]$, $[-0.5,0.5]$
        & $[0.5,1.5]$, $[-2.0,2.0]$ \\
        \bottomrule
\end{tabular}
\end{table}

\FloatBarrier
\section{Baseline Comparison Details}
\label{app:baseline_comparison_details}

All baselines use the same fixed-OOD evaluation suite and success criterion as Table~\ref{tab:core_domain_results}. We do not claim strict wall-clock equalization across different training codebases; instead, Table~\ref{tab:core_domain_results} fixes the evaluation protocol, while Figure~\ref{fig:sample_efficiency} reports the cost-normalized view.

\begin{figure}[hbt]
    \centering
    \includegraphics[width=0.50\linewidth]{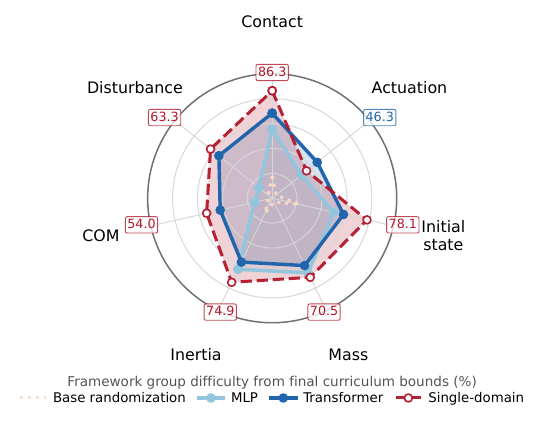}
    \caption{Single-domain HORIZON policies reach larger mastered ranges than the full multi-domain run on their corresponding axes.}
    \label{fig:single_domain_expert_reachability}
\end{figure}

\textbf{Multi-expert distillation.} The teachers are HORIZON-trained single-domain policies selected from mastered checkpoints. Figure~\ref{fig:single_domain_expert_reachability} shows that these teachers reach larger mastered ranges than the full multi-domain HORIZON run on their own domain axes, so the student is not limited by weak nominal teachers. The distilled student uses the same deployment actor interface as HORIZON and receives no domain identifier at test time. Its lower composed OOD result therefore indicates that compressing isolated single-axis behaviors does not replace joint on-policy training under simultaneous perturbations.

\textbf{GRAM.} GRAM is evaluated on the same fixed-OOD suite, but it keeps its method-specific actor and adaptation architecture rather than using the HORIZON actor exactly. We adapt the observation and evaluation interface to the Go2 task and the same OOD knobs. The GRAM run uses its adapted training setup without an exhaustive hyperparameter sweep, and serves as an adaptation baseline for comparing HORIZON against a method with a different robust-control architecture under the same evaluation protocol.

\textbf{ADR.} ADR uses the same locomotion training stack, reward, and evaluation interface where possible, with the main difference being adaptive randomization over the physical-domain groups. PPO and actor settings follow the training configuration in Table~\ref{tab:ppo_settings}, while ADR scheduler parameters are fixed before final evaluation rather than tuned against the reported OOD table. This baseline tests whether adaptive distribution widening alone can handle the composed seven-domain stress.

\section{Real-World Hardware Details}
\label{app:real_world_hardware_details}

Table~\ref{tab:real_world_hardware_settings} summarizes the physical interventions used in the hardware evaluation. Each condition is executed zero-shot by both HORIZON and the base-DR baseline with the same deployed actor interface; no hardware-specific fine-tuning or privileged state input is added. We treat these trials as difficult deployment tasks: after the mechanical intervention is applied, a run is considered successful only if the robot can remain balanced and sustain commanded locomotion without external support.

\begin{table}[hbt]
    \centering
    \setlength{\tabcolsep}{3pt}
    \caption{Hardware settings for the real-world evaluation.}
    \label{tab:real_world_hardware_settings}
    \begin{tabular}{@{}p{0.26\linewidth}p{0.66\linewidth}@{}}
        \toprule
        Test condition & Hardware setting \\
        \midrule
        Front-knee locked
        & Front knees fixed at the Go2 initialization-pose minimum angle. \\
        Hind-knee locked
        & Hind knees fixed at the Go2 initialization-pose maximum angle. \\
        Stilt / tied-leg walking
        & Calves extended by 15\,cm on all legs. \\
        Hind-leg-only stilts
        & Calves extended by 15\,cm only on the hind legs. \\
        Dynamic payload
        & 3\,kg or 5\,kg iron ball attached to the body. \\
        Combined two-ball load
        & Two iron-ball payloads attached simultaneously. \\
        High-load stilted locomotion
        & Calves extended by 15\,cm, combined with the payload setting. \\
        \bottomrule
    \end{tabular}
\end{table}

The base-DR baseline exposes the limits of near-nominal randomization under these perturbations. It cannot produce forward locomotion in the tied-leg/stilted-leg tests, and it fails to balance when the front knees are locked. With only the hind knees locked, it can barely remain upright but does not produce reliable locomotion. Under dynamic payload, it can marginally preserve balance with the smaller 3\,kg ball, while heavier or coupled payload settings further destabilize the controller.

\FloatBarrier
\section{Cross-Robot Simulation Control}
\label{app:cross_robot_control}

\begin{figure}[hbt]
    \centering
    \includegraphics[width=1.0\linewidth]{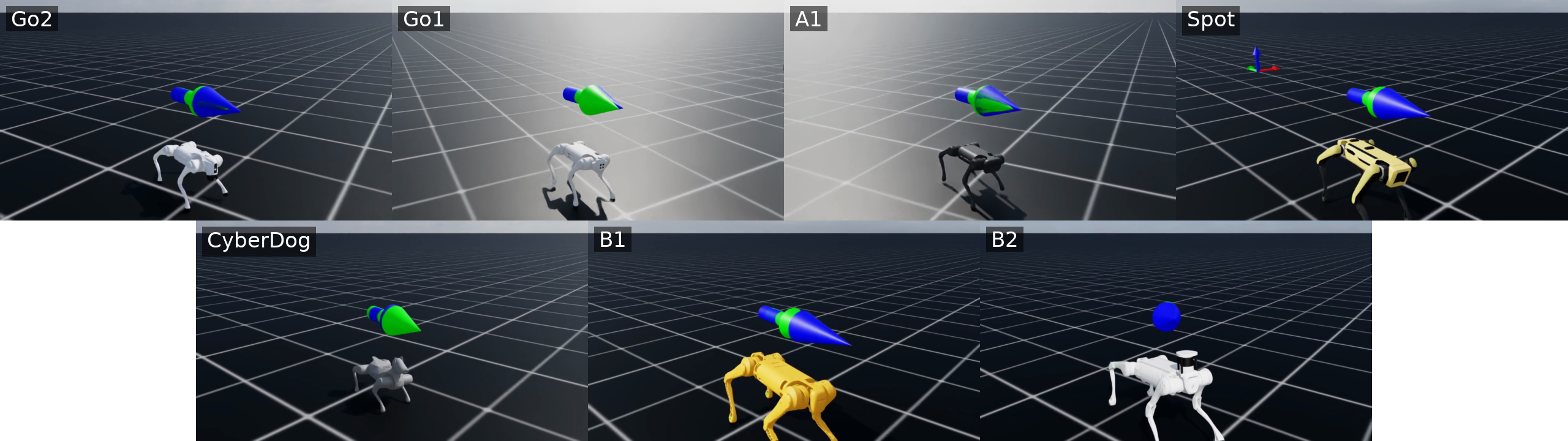}
    \caption{Representative cross-robot simulation control frames. The montage includes Go2, Go1, A1, Spot, CyberDog, B1, and B2 rollouts under base locomotion commands.}
    \label{fig:cross_robot_control_montage}
\end{figure}

Beyond the Go2-centered physical perturbation tests, we also ran simulation rollouts on additional quadruped morphologies using the same deployment-style control interface. Figure~\ref{fig:cross_robot_control_montage} shows representative frames from these cross-robot base-command checks. These trials provide supplementary evidence of morphology coverage and are not used in the fixed-OOD success tables.

\section{Core-Selection Check}
\label{app:heldout_core_selection}

The pairwise composition matrix in Fig.~\ref{fig:domain_composition} selects the compact domain core, while Table~\ref{tab:core_domain_results} evaluates the same fixed-OOD family. To check for possible selection leakage, we add frozen-checkpoint evaluations on held-out and core-extension compositions. No policy is retrained and no curriculum frontier is changed.

We evaluate independently trained core-domain and full-domain checkpoints with the same fixed-eval code path, using $1024$ parallel environments and one episode per environment for each of five training seeds. The first six rows are held-out compositions not used to choose the actuation, mass, disturbance, and COM core; the last four rows test the selected core and one-domain extensions.

\begin{table}[hbt]
    \centering
    \setlength{\tabcolsep}{3pt}
    \caption{Held-out and core-extension results. Success uses the same fixed-eval criterion as Table~\ref{tab:core_domain_results} and is reported as percentage points over five independently trained seeds.}
    \label{tab:heldout_core_results}
    \begin{tabular}{@{}p{0.46\linewidth}p{0.15\linewidth}p{0.15\linewidth}p{0.10\linewidth}@{}}
        \toprule
        Active OOD groups & Core & Full & Diff. \\
        \midrule
        Contact + inertia + initial state & $89.7 \pm 0.7$ & $94.8 \pm 0.2$ & $-5.1$ \\
        Actuation + mass + contact + initial state & $71.7 \pm 1.0$ & $72.3 \pm 1.0$ & $-0.6$ \\
        Actuation + disturbance + inertia + initial state & $45.9 \pm 1.1$ & $43.5 \pm 1.4$ & $+2.4$ \\
        Mass + COM + contact + inertia & $54.4 \pm 0.7$ & $57.2 \pm 1.9$ & $-2.8$ \\
        Actuation + mass + inertia + contact & $73.2 \pm 0.7$ & $72.6 \pm 0.9$ & $+0.6$ \\
        Disturbance + COM + contact + initial state & $53.8 \pm 1.0$ & $49.3 \pm 1.4$ & $+4.5$ \\
        Actuation + mass + disturbance + COM & $47.2 \pm 1.6$ & $37.9 \pm 0.8$ & $+9.3$ \\
        Selected core + contact & $46.8 \pm 1.4$ & $34.2 \pm 0.9$ & $+12.6$ \\
        Selected core + inertia & $43.2 \pm 1.2$ & $33.2 \pm 0.7$ & $+10.0$ \\
        Selected core + initial state & $46.1 \pm 1.4$ & $35.9 \pm 0.4$ & $+10.2$ \\
        \midrule
        Average & $57.2$ & $53.1$ & $+4.1$ \\
        \bottomrule
    \end{tabular}
\end{table}

The held-out rows are mixed: the compact core wins seven of ten unseen compositions, while full-domain training is stronger on non-core-heavy and contact-heavy rows. By contrast, the selected core and all one-domain extensions outperform full-domain training, with an average gain of about ten percentage points across the four extension rows. Thus Table~\ref{tab:core_domain_results} supports a recoverable compact core for the target composed stress and nearby extensions, but not a universal core optimum across all domain compositions.

\end{document}